\documentclass[journal]{IEEEtran}

\ifCLASSINFOpdf
   \usepackage[pdftex]{graphicx}
\else
\fi

\begin{document}

% \title{Neural Trees Growing in the Head: from Extracting Hierarchical Structures in the World to Shaping the Brain Itself}
%\title{Brain Developmental Hypothesis on the Frontal Cortex to Extract Ordinal Structures in the World and Shape its Organization Itself}
% \title{Brain Developmental Model from Extracting Ordinal Structures in the World to Top-Down Memory Organization}
% \title{Complementary Developmental Model from Extracting Order in the World to Ordering the Brain}
%\title{Coding Ordinal Information in Neurons: from Extracting Structure in the World to Self-Organizing the Brain}
% \title{Coding Ordinal Information in Frontal Neurons: Developmental Model for Extracting Structure to Self-Organizing the Brain}
% \title{Coding Structure in Frontal Neurons: Developmental Model for Extracting Order in Signals to Brain Self-Organization}
%\title{Serial Order in Frontal Neurons: Models Review from Brain Development to Higher-Level Cognition}
\title{A Review of Models for Serial Order in Frontal Neurons: from Brain Development to Higher-Level Cognition}
% \title{Frontal Neurons to Code Ordinal Information: from Extracting Structure in the World to Self-Organizing the Brain}
% in the complementary learning systems framework
% \author{Alexandre~Pitti, %~\IEEEmembership{Member,~IEEE,}
%         Mathias~Quoy, %~\IEEEmembership{Fellow,~OSA,}
%         Catherine~Lavandier, %~\IEEEmembership{Fellow,~OSA,}
%         Sofiane~Boucenna, %~\IEEEmembership{Life~Fellow,~IEEE}% <-this % stops a space
%         Wassim~Swaileh~%~\IEEEmembership{Life~Fellow,~IEEE}% <-this % stops a space
%         and~Claudio~Weidmann%~\IEEEmembership{Life~Fellow,~IEEE}% <-this % stops a space
% \IEEEcompsocitemizethanks{\IEEEcompsocthanksitem Authors are with ETIS Lab, UMR8051, CY Cergy Paris University, ENSEA, CNRS, Cergy-Pontoise, France \protect\\ e-mail: alexandre.pitti@cyu.fr.}% <-this % stops a space
% \thanks{Manuscript received April 19, 2005; revised August 26, 2015.}}

\author{Alexandre~Pitti, %~\IEEEmembership{Member,~IEEE,}
        Mathias~Quoy, %~\IEEEmembership{Fellow,~OSA,}
        Catherine~Lavandier, %~\IEEEmembership{Fellow,~OSA,}
        Sofiane~Boucenna, %~\IEEEmembership{Life~Fellow,~IEEE}% <-this % stops a space
        Wassim~Swaileh~%~\IEEEmembership{Life~Fellow,~IEEE}% <-this % stops a space
        and~Claudio~Weidmann%~\IEEEmembership{Life~Fellow,~IEEE}% <-this % stops a space
\thanks{Authors are with ETIS Lab, UMR8051, CY Cergy Paris University, ENSEA, CNRS, Cergy-Pontoise, France \protect\\ e-mail: alexandre.pitti@cyu.fr}% <-this % stops a space
\thanks{Manuscript received April 19, 2005; revised August 26, 2015.}}

% The paper headers
\markboth{Journal of \LaTeX\ Class Files,~Vol.~14, No.~8, August~2015}%
{Shell \MakeLowercase{\textit{et al.}}: Bare Demo of IEEEtran.cls for IEEE Journals}

\maketitle

% The paper headers
% \markboth{Journal of \LaTeX\ Class Files,~Vol.~14, No.~8, August~2015}%
% {Shell \MakeLowercase{\textit{et al.}}: Bare Demo of IEEEtran.cls for Computer Society Journals}

%\IEEEtitleabstractindextext{%
\begin{abstract}
In order to keep trace of information and grow up, the infant brain has to resolve the problem about where old information is located and how to index new ones. We propose that the immature prefrontal cortex (PFC) use its primary functionality of detecting hierarchical patterns in temporal signals as a second purpose to organize the spatial ordering of the cortical networks in the developing brain itself. 
Our hypothesis is that the PFC detects the hierarchical structure in temporal sequences in the form of ordinal patterns and use them to index information hierarchically in different parts of the brain. % as neural trees. 
Henceforth, we propose that this mechanism for detecting patterns participates in the ordinal organization development of the brain itself; i.e., the bootstrapping of the connectome. 
By doing so, it gives the tools to the language-ready brain for manipulating abstract knowledge and planning temporally ordered information; i.e., the emergence of symbolic thinking and language.
%
%We will present two experiments showing how our architecture can index hierarchically information in the text domain and how it can model the development of complex and hierarchically organized neural networks.
We will review neural models that can support such mechanisms and propose new ones. We will confront then our ideas with evidence from developmental, behavioral and brain results and make some hypotheses, for instance, on the construction of the mirror neuron system, on embodied cognition, and on the capacity of learning-to-learn. 

\end{abstract}

%}

% \maketitle

%\IEEEdisplaynontitleabstractindextext

\IEEEpeerreviewmaketitle

%\IEEEraisesectionheading{\section{Introduction}\label{sec:introduction}}
\section{Introduction}\label{sec:introduction}

More than any other area in the brain, the development of the prefrontal cortex (PFC) early in infancy plays a major role for the acquisition of models, patterns and for the manipulation of structured knowledge. 
For instance, the features of the PFC play an important role for developing logical inference and algebra, for the acquisition of language and music~\cite{Dehaene2015, Koechlin2014, Koechlin2016, Koechlin2018}, for the learning of task sets and the resolution of rule-based problems~\cite{Romo1999, Tanji2001, Wang2018}.
We propose that what does essentially the PFC is to manipulate {\it items} and {\it patterns} independently; where a pattern is a sequence, a cluster or a group of several items or units. 
Using this tool, we think the PFC participates altogether to the development of abstract thinking and to the hierarchical organization of the brain. 

Observations in the Broca area and in the pre-Supplementary Motor Area (pre-SMA) located in the PFC confirmed the existence of such neurons sensitive to ordinal and hierarchical patterns only. For instance, some of them were found sensitive to the temporal order in audio sequences and to proto-grammars (i.e., a rudimentary grammar) but not to the particular sound emitted~\cite{Friederici2011, Gervain2008, Gervain2017}. 
Other neurons were found sensitive to the different levels of complexity and depth in sequences~\cite{Petersson2004, Fitch2012}.
In the motor domain, a majority of neurons in PFC were found sensitive to the temporal patterns of motor behavior such AABB or ABAB and a minority to the particular motor units~\cite{Tanji2001, Shima2007, Tanji2007}. Moreover, different ones were observed salient to the temporal coherence in visual scenes~\cite{Fadiga2009}; e.g. its semantic. Some similar results were found with neurons sensitive to orders, schemata in spatial contexts~\cite{Barone1989}, and to geometrical rules in the recognition of shapes~\cite{Averbeck2003a, Averbeck2003b} or in visual sequences~\cite{Wang2019}.
Surprisingly, these neurons were all found insensitive to the particular sound, action or visual information composing the sequence presented \emph{per se}. %, but only to the specific orderly patterns, hierarchical structure or schemata that they were encoding such as AABB or ABAB or AAAA, or to a relative order in a temporal sequence (e.g., the beginning, the second place or at the end) or to a relative location in space. 

% FIG 1 Separating Temporal Code and Neurons Index
\begin{figure*}[t!]
\begin{center}
\includegraphics[width=16cm]{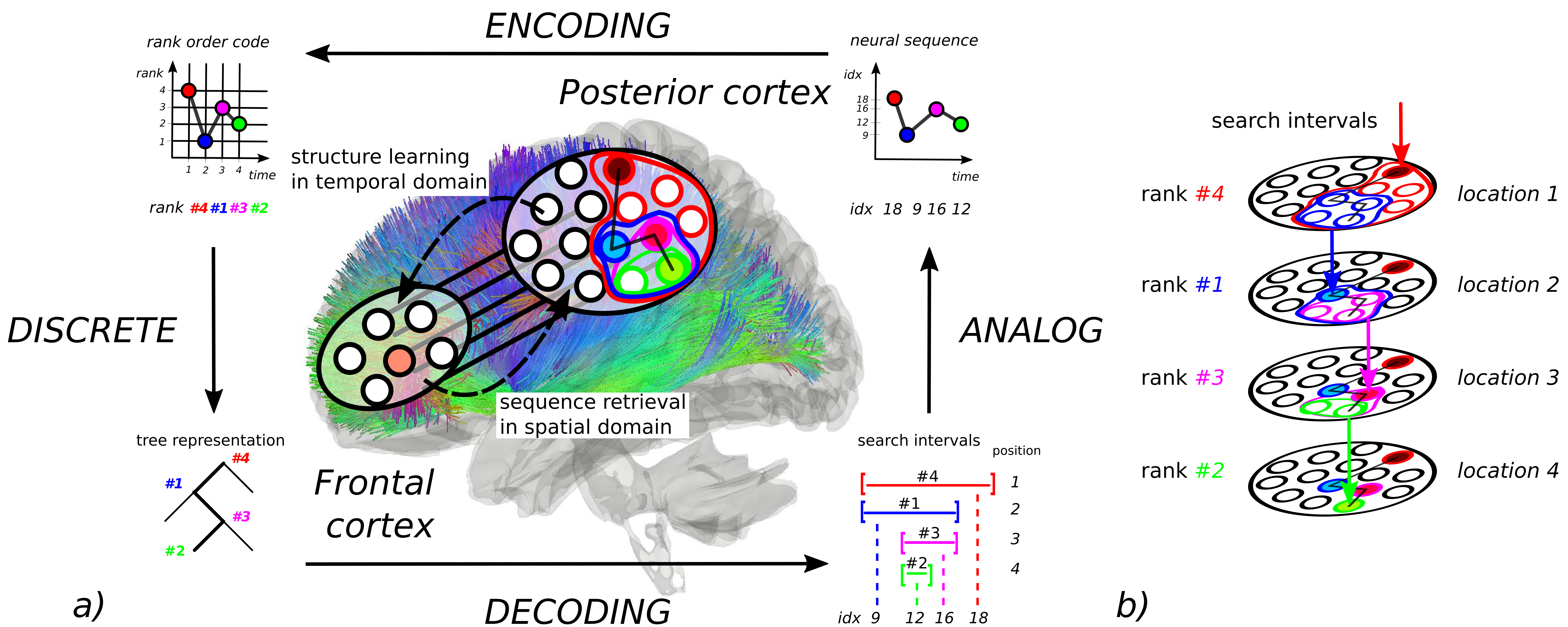}
\end{center}
\vspace{0.25cm}
\begin{center}
\includegraphics[width=16cm]{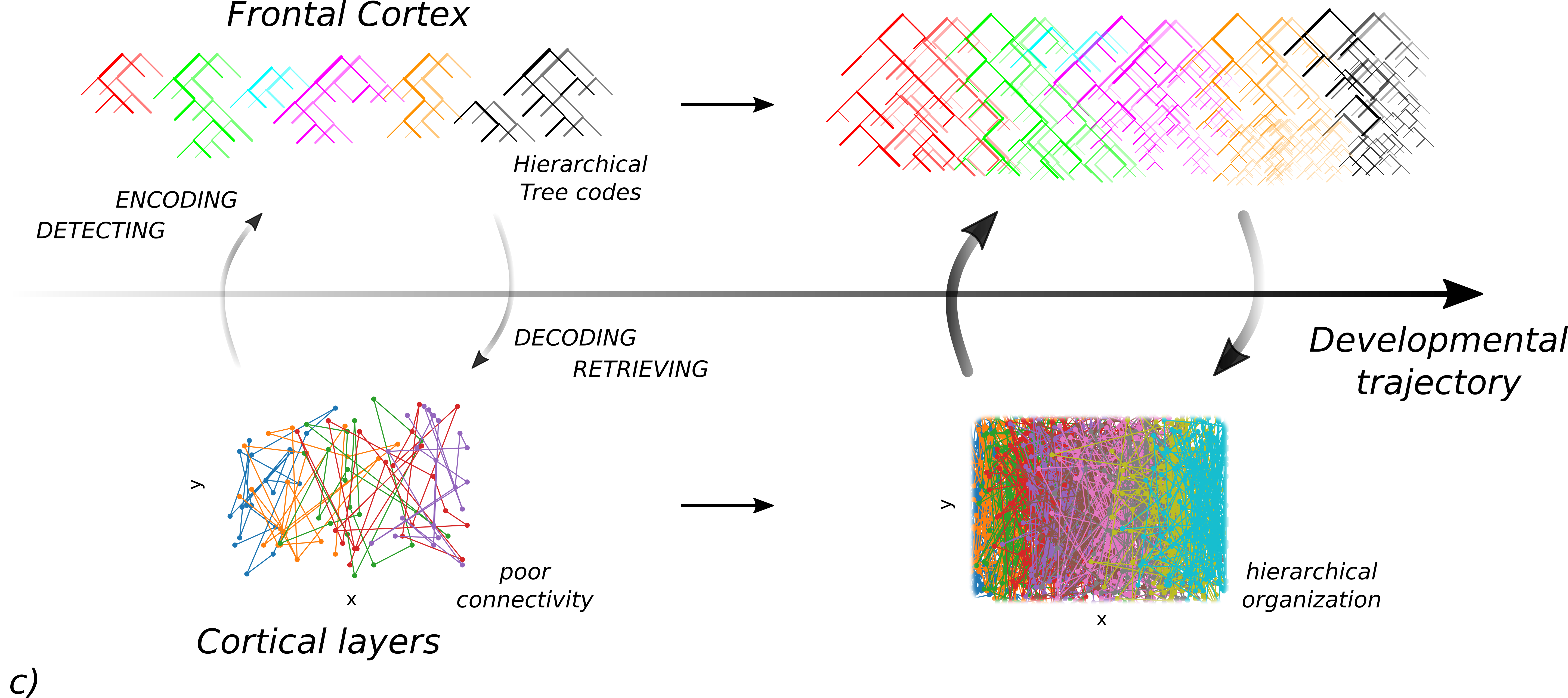}
\end{center}
\caption{
Global schema on our proposal for the brain development for acquisition of hierarchical and semantic knowledge and functional organization of memory. In a), we propose that the conjunctive cells in the Prefrontal Cortex learn and recognize the order and structure in temporal sequences as trees or rank codes. They use these codes to organize spatially the location of clusters and memories. 
 In b), these trees and rank codes can serve to retrieve and decode the memory location in a top-down manner. 
In c), we suggest that during the infant development these two processes are at work. The frontal areas of the brain bootstraps and co-learns with its posterior parts the ordinal and hierarchical structure in temporal sequences from the outside world to organize spatially and hierarchically the memories in brain networks.
Ordinal and tree codes participate to what we call the self-structuring of the brain's organization itself and to acquisition of a neuro-symbolic representation of information. 
~\emph{Image Source: http://brain.labsolver.org/,~\cite{Yeh2018}.}
}
\label{fig:connectome}
\end{figure*}

% 1. le mécanisme de base
% NEW
One core proposal is that the ability of the PFC to extract and represent the hierarchical structure out of raw data has served for the self-structuration of the brain itself. 
% NEW
Although this proposal is certainly not novel~\cite{McClelland2003, Buschman2014}, the implementation of it is still subject to many hypothesis~\cite{Badre2018} and we think that ordinal neurons can provide some computational features for modeling it.
We will present several models of neurons that can encode ordinal and hierarchical information in their synaptic weights and represent the item's order in sequences; i.e., the elements' {\it hierarchy} or  {\it rank} with respect to the others~\cite{Botvinick2007, Pitti2020}. %Some of them have been tested successfully in previous papers~\cite{Botvinick2007, Pitti2020}, they model in different ways the same function.
%We propose that spiking neurons possess all the features for encoding ordinal and hierarchical information in their synaptic weights to represent the item's order in sequences; i.e., the elements' {\it hierarchy} or  {\it rank} with respect to others. 
% NEW
In our view, these implementations can be derived from the learning mechanism of spike timing-dependent plasticity (STDP) or from the rank-order coding algorithm~\cite{Thorpe2001} to code ordinal patterns as rank-order codes. Actually, they can be seen as hierarchical neural trees as they encode ordinal patterns. %, although their biological plausibility of it can be discussed.
%Such implementation can be derived from the learning mechanism of spike timing-dependent plasticity or from the rank-order coding algorithm~\cite{Thorpe2001} to code ordered patterns as rank-order codes. From our point of view, this implementation can be seen as hierarchical neural trees.
%NEW
We will discuss as well some other neural architectures that have realized the same function.%, such as push-up/push-down working memories and gated neural networks~\cite{Beiser1998, Botvinick2006, Rougier2002, Trenton2013, Hasselmo2018, Wang2018}, and reservoir computing~\cite{Enel2016}.

% 2. H1
% NEW
Based on this idea, 
our first assumption or hypthosesis (H1), also shared by others, is that the conjunctive cells in PFC have evolved as sort of serial order detectors in temporal sequences; see Fig.~\ref{fig:connectome} a). % , but not what type of information is encoded (the neurons' value or item they represent \emph{per se}) \emph{per se}  for retrieving back where information is located (the neurons' address).
%NEW
In some other works, they are considered as neuronal pointers or timestamp to abstract information~\cite{Zylberberg2011, Eliasmith2012}. 
%In other words, the prefrontal cortex may have learned neuronal pointers to retrieve back information~\cite{Zylberberg2011, Eliasmith2012}. 
%
%As a metaphor, we can see the PFC to act as a mailman, having no idea about the identities of the sender or of the receiver but an exact knowledge where they are located (e.g., their mailing address).
% 
%Therefore, in order to drive robustly the retrieval of one durable information, the PFC must use these conjunctive cells to \emph{encode} redundantly its location (e.g., through population coding).

% 3. H2
Our second hypothesis (H2) is that these conjunctive cells, along with the dense long-range axons -- serving to connect the PFC to other parts of the brain,-- are exploited to retrieve back the location of neurons in posterior brain areas. Their 
second purpose is for selecting separated items from different parts of the brain but related into one cluster, for broadcasting and routing information at the brain level, see Fig.~\ref{fig:connectome} b). This second hypothesis is still debated. %less intituive and less accepted, but not new.

% 4. H3 NEW
In a third hypothesis (H3), we defend the idea that these two mechanisms for encoding the ordinal structure in the world (H1) and for decoding their neural sites (H2) are working along, side-by-side, toward the self-structuring of the brain's organization through a developmental pathway; see Fig.~\ref{fig:connectome} c). This third developmental hypothesis is to our knowledge novel in the literature.

%The better the PFC's neurons to detect ordinal patterns in temporal sequences, the better they organize the spatial topology of brain networks into hierarchies, and the more efficient the information processing is done to bind and retrieve flexibly related but unconnected elements together; i.e., chunking and unchunking elements into a sequence.
First, the well-ordering of the neocortex into hierarchies and small-world dynamics serve for fast and robust information retrieval of codes, which is not something clearly acknowledged in the computational neuroscience literature and no mechanism for cluster retrieval in complex systems is proposed in that sense. 
For instance, in most researches, only the topological aspect of complex networks is emphasized such as the balanced level between segregation and integration~\cite{Varela2001, Sporns2004, Tognoli2014}, but the way how information is then retrieved in those networks is not examined.
 Second, we propose that the mechanism for structure learning and decoding in the PFC may serve, as a second feature, to shape the well-ordering of the other parts of the brain for efficacy (H2). 
%
% Detecting one structure in sequences gives the opportunity to generate sequences that possess also this particular structure, in an unsupervised manner.
%
Hence, we propose that the mechanisms for extracting structure in the world, performed by the PFC, serve for structuring memory networks in the brain itself, in line with the ideas of self-structuring information~\cite{Lungarella2005, Byrge2014}, neural scaffolding~\cite{Changeux1989a} and embrainment~\cite{McClelland2010}, see Fig.~\ref{fig:connectome} a-c). 

Altogether, we think that these three mechanisms bootstrap the brain to gain the capability to process hierarchical information and to access higher-level skills necessary for problem-solving tasks, for the production of rule-based behaviors and for the flexible planning of actions~\cite{Greenfield1991, Koechlin2006b, Buschman2014, Dehaene2015, Varona2016, Badre2018}. 
They enable the brain to create abstract models and syntactic rules in order to generate new sequences built from atomic components (e.g. motor units, visual elements, spoken words, etc.).
We suggest further that this processing has served for robust memory recall and avoidance of catastrophic forgetting, for faster processing and for energy efficiency.

As a twist of evolution and of embodiment, we propose that these neural mechanisms, whose purpose served to detect the order in incoming signals, could have shaped as well the brain order, its functional organization and its cognitive development, leading path to the language-ready brain and to the symbolic mind.

% NEW
We will first present 
several neural architectures from the litterature for coding the serial order in sequences.  We will comment the pros and cons of each, and their biological plausibility associated to their computational advantages. % in section~\ref{StateofArt}.
In section~\ref{Model}, we present our model for coding the ordinal and hierarchical structure in sequences. 
%In section~\ref{sec:experiments}, we will present then one experiment for encoding and generating a large text corpus based on the encoding of structural information found in sentences.
%
%In a second experiment, we will show how ordinal rank and tree codes can generate complex and hierarchically-ordered networks with similar properties with small-world networks, although designed with different rules.
%
In section~\ref{Discussion}, we will draw then links with neurosciences, cognitive sciences and developmental sciences.

\section{Neural architectures for serial order in sequences}
\label{StateofArt}

We present in this section several architectures used for detecting structure and ordering items in sequences based on various mechanisms. 

\subsection{Task-setting Networks} % Rigotti, Botvinick2006, O'Reilly, Beiser and Houk

A majority of neural architectures exploits attentional mechanisms and priority signals for modeling serial processing, see Fig.~\ref{fig:conjunctive_rank_order} e).
Control nodes~\cite{Miller2001} or contextual neurons have been employed by \cite{Beiser1998, Botvinick2006} to hold on and release information for the serial processing of events in models of the PFC.
A similar concept has been used by Tani with parameter bias neurons~\cite{Tani2004} in recurrent neural networks to detect one context or to switch to another one in robotic tasks for grasping objects.

Besides, Dehaene and colleagues~\cite{Zylberberg2010, Zylberberg2011} propose the use of task-set neurons to prioritize or to hold on sensory events with respect to others.
The architecture is constituted of inter-connected spiking neurons that route differently the information with respect to the task.
Such system is capable to hold on one activity signal a certain amount of time till the release of a second signal.
The detection of erronous rules can be also treated with the use of predictive coding accounting for the mismatch negativity~\cite{Wacongne2012}.

Such strategy has been used also by O'Reilly and colleagues~\cite{Rougier2002, Reilly2006, Trenton2013} and others~\cite{Abrossimoff2020, Granato2020} with the control of up-state and down-state with neuromodulatory signals to solve several cognitive tasks like the Wisconsin card game and task-switching.

Attention mechanisms have been used successfully in the popular Long Short-Term Memory networks (LSTM) and gated networks in order to achieve long-term dependencies~\cite{Hochreiter1997, Gers2000}.
Several subnetworks are used to learn, forgot, or hold on signals replacing the functions of neuromodulators. % in the models of working memory cited above.
Wang and colleagues~\cite{Wang2018} use a variant of the LSTM to solve the task-set problem of Harlow~\cite{Harlow1949} for learning-to-learn and for which standard reinforcement learning algorithms are inefficient. After many trials, the network becomes sensitive to the strategy to adopt in the first exploration steps and independently to the items' value.
%
%We will provide a scenario in section~\ref{sec:learning-to-learn} to explain how Harlow's task-set problem could be solved with an ordinal strategy.

\subsection{Hierarchical Tree Networks} % Pitti2020, Badre2018, Dehaene, Pezzulo, Botvinick
%Botvinick MM. Hierarchical models of behavior and prefrontal function. Trends in Cognitive Sciences, 12(5): 201e208, 2008.

Although it has been suggested that the brain do not represent information as hierarchical tree structures, but more like Markovian processes~\cite{Frank2012}, this argument is still debated. Many reseachers emphasize the importance of hierarchical tree structures for learning information at different state abstraction and for modeling higher-level cognition~\cite{Frank2012, Fitch2012, Dehaene2015, Badre2018}, see Fig.~\ref{fig:conjunctive_rank_order} c).

In order to implement conditional rules, tree-like algorithms are conceptually useful for modeling branching, recursivity and serial order in sequences. These features are known difficult to model in standard neural networks in comparison to classic AI.
Therefore, models that mimic the tree search algorithms in classic AI have been proposed for implementing the fast sorting, the planning of information, conditional branching between sub-tasks, which are an important feature of working memory~\cite{Buschman2014, Ranti2015, Maisto2021, Badre2018}.

Ordinal codes are interesting in this part, because they avoid the use of tree models as there is an equivalence between ordinal permutations, stack-order codes and ordinal tree codes. 
In section~\ref{Model}, we provide two ways to implement tree-like hierarchical structure with ordinal codes and spikes with a Huffman-like ordinal codes and an ordinal STDP rule.

\subsection{Reservoir Computing} % Dehaene, Dominey, Rougier
Recurrent neural networks (RNN) have been widely used for learning temporal sequences \cite{Dauce2002}. Reservoir computing is a framework for learning temporal dynamics where a reservoir of neurons (a RNN) express different dynamics under external inputs, see Fig.~\ref{fig:conjunctive_rank_order} d). A readout mechanism transforms the nonlinear high dimensional dynamics into output neurons. In particular EchoState Networks (ESN) have a recurrent neural network (reservoir) and a readout layer with adaptable weights \cite{Jaeger2001}. Liquid state machines use the same conceptual framework with spiking neurons \cite{Maas2002}. A complete review may be found in \cite{Lukosevisius2009}, and a recent link between neural substrates of ESN in \cite{Dominey2021}. ESN have been used in various applications, and in particular in sequence learning for motor control \cite{Laje2013} and grammar learning for speech processing \cite{Dominey2006,Hinaut2013}. \cite{Enel2016} describes how nonlinear properties can be achieved in those networks like the learning of serial rules. 
One drawback of ESN is that learning is performed offline using a large training dataset. %Our INFERNO model enables an online adaptation \cite{}.
As acknowledged in~\cite{Enel2016}, the underlying mechanism of reservoir computing behind nonlinear-mixed selectivity and the processing of serial order has still to be better explained and understood.

\subsection{Gain-Modulatory Networks} % Hasselmo2018, Donnarumma2015, 

\begin{figure*}[t!]
\begin{center}
\includegraphics[width=16cm]{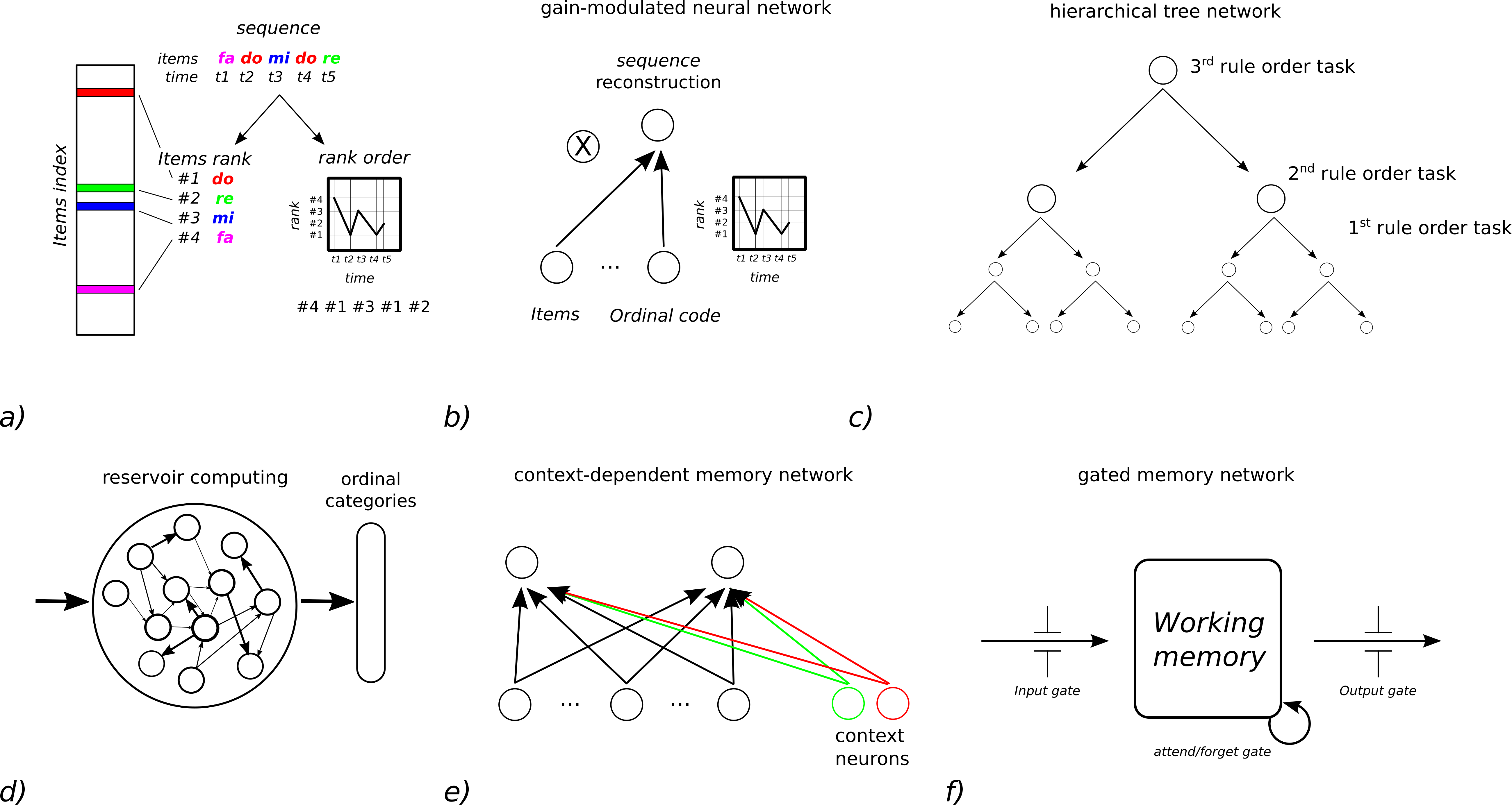}
\end{center}
\caption{Neural architectures for serial order coding in working memory. In a-b), neural architectures with gain-modulatory neurons sensitive to ordinal information in sequences can extract an ordinal pattern and can combine it with new items into a new sequence~\cite{Botvinick2007}. c) the conditional branching of hierarchical trees can direct the sequences reconstruction based on their importance in the tree~\cite{Ranti2015, Badre2018}. d) reservoir networks can learn to generalize the serial order in sequences and then create output neurons sensitive to the items location in the sequence~\cite{Enel2016}. e) neural architectures with contextual neurons or task-set neurons can modulate their output activity with respect to a task rule~\cite{Miller2001}. Gated networks are a variant of them as they have special signals to gate, maintain or forget signals with respect to the current task~\cite{Botvinick2006, Wang2018, Abrossimoff2020, Granato2020}.}
\label{fig:conjunctive_rank_order}
% \end{figure*}
% \begin{figure*}[t!]
\begin{center}
 \includegraphics[width=16cm]{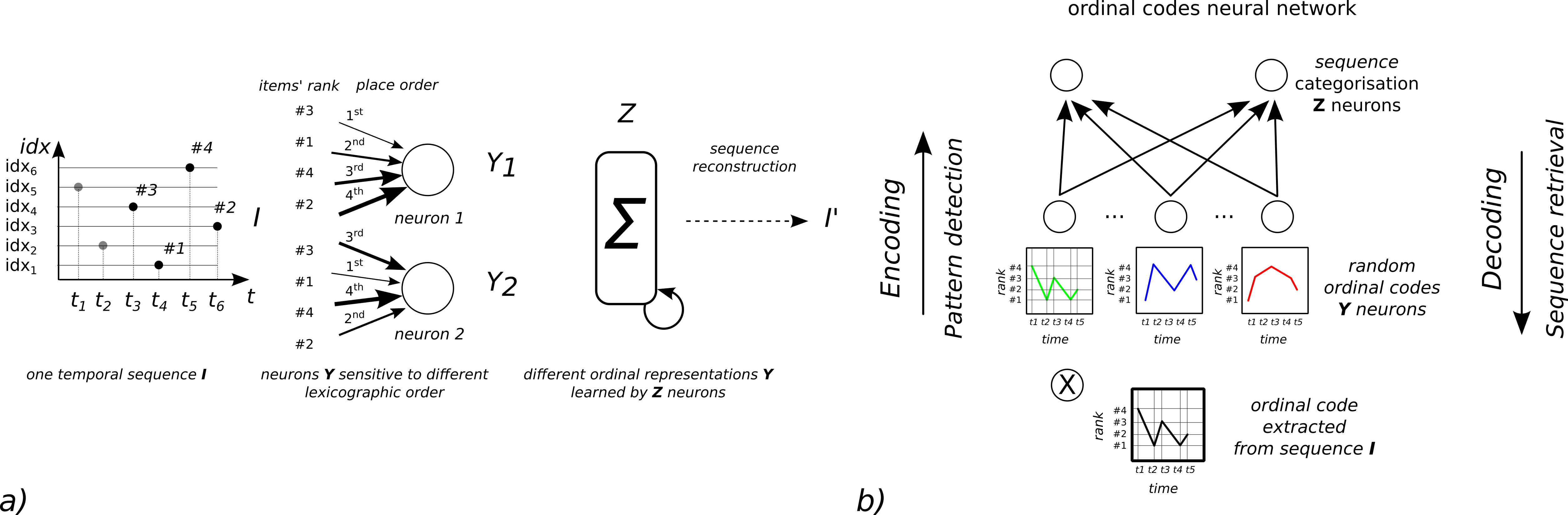}
\end{center}
\caption{Proposed architecture for coding the serial order in sequences. In a), the gating operation for encoding structure in sequences is based on the relative order of each element in the sequence $I$, and with different lexicographic orders for which each neuron $Y_1$ and $Y_2$ are sensitive to. The neurons $Z$ are sensitive to the $Y$ population code activity~\cite{Pitti2020}. In b), the neural population $Z$ serves to retrieve one sequence $I'$ close to the original sequence $I$ by error minimization (predictive coding).}
\label{fig:conjunctive_rank_order2}
\end{figure*}

Conjunctive cells are neurons that bind mutual information together by gain-modulation. The result is in the form of a multiplicative  effect across two or more signals, which is useful for updating online one output signal \emph{conditionally} with one entry without updating the synaptic weights, see Fig.\ref{fig:conjunctive_rank_order} b). This allows conditional reasoning with sub-branches without the need of modeling hierarchical trees.
These cells in the prefrontal cortex permit to represent conjunctive representations~\cite{Genovesio2014} such as spatial or temporal order and crossing items with distance or duration information~\cite{Barone1989, Inoue2006}. 

Botvinick and Watanabe propose a gain-field network to bind items with a temporal order template in a model of the PFC in~\cite{Botvinick2007}. %The idea is similar to our model presented in section~\ref{Model} expect that we extract the ordinal template from the input sequences instead of binding items to the template for constructing the sequence, see Fig.~\ref{fig:conjunctive_rank_order} a). %They could replicate the experimental results from XXX for ordinal sequences.
Differently, Donnarumma and colleagues follow a reverse coding strategy by extracting the structural information in the shape of a multiplicative network called a 'programmer' in \cite{Donnarumma2015}. By instantiating different variables, the programmer network can use the expression-free syntax learned and create new output for the cognitive task called 1-2-A-X. A similar idea has been proposed by~\cite{Tanji2007} to describe how the PFC can extract or generalize structural information in action sequences.

Besides, Hasselmo and Stern describe another multiplicative network, which was used for binding multiple variables of different categories, to change the rules dynamically without modifying the synaptic weights in~\cite{Hasselmo2018}. A similar gain-field model has been proposed to solve the Wisconsin test~\cite{Tanji2001}.

These multiplicative networks have for advantages to respond conditionally to variable inputs, and to detect and exploit regular expressions. These features are useful to avoid the conditional sub-branches of recursive systems difficult to implement. 

Our model based on ordinal codes, published in~\cite{Pitti2020} and described in section~\ref{Model}, exploits several features of these different architectures, see Fig.~\ref{fig:conjunctive_rank_order2}. 
For instance, %in order ordinal code similar based on multiplicative networks. They exploit the 
the gating operation for encoding structure in sequences is based on the relative order of each element in the sequence $I$. This part is similar with the gain-modulatory mechanism that models the conjunctive cells to extract or bind a temporal pattern with items.
In second, a multiplication is done between this ordinal pattern and the different lexicographic orders learned in the neural population $Y$ for which each neuron is sensitive with.
These neurons encode random ordinal patterns to represent nonlinearly the incoming sequence. This part is similar with reservoir computing.
Finally, the neural population $Z$ serves to retrieve one sequence $I'$ close to the original sequence $I$ by error minimization and predictive coding. Similar with task-based networks, ours selects, maintains and switches context to generate sequences of items that does not violate the ordinal rules followed by $Z$ neurons.

\section{Neural implementation of ordinal codes}
\label{Model}

\subsection{Computational features of ordinal codes}
\label{sec:roc_features}

Ordinal and tree codes can be implemented to learn the underlying hierarchical structure within a sequence. This can be done by weighting the relative order of the elements within a sequence depending on their rank or their relative importance (i.e. their hierarchical importance or depth-level). Hence, this type of coding differs from the temporal order codes done in Hebbian learning, conventional recurrent neural networks and bio-inspired Spike Timing-Dependent Plasticity reinforcement rule.

% NEW
We can use the example presented in Fig.~\ref{fig:tree_code} to explain the difference between these three types of coding; i.e., STDP, ordinal coding and tree-order codes. We add as well an implementation of ordinal codes, more biologically plausible, based on hebbian learning for associative and recurrent networks (i.e., Hopfield-like networks and spiking networks).
For instance, in a time series of six elements ordered as follows $seq:[18, 13, 8, 14, 5, 19]$, see Fig.~\ref{fig:tree_code} a) and b), the rank code corresponding to the relative order of each element with respect to others is $rank:[\#5, \#3, \#2, \#4, \#1, \#6]$. This coding can be seen as an analog-to-digital conversion as the index of each element are no more present in the sequence. This drastic quantization of information represents a computational advantage as the ordinal code is now reduced to $N$ elements to store and not $M$, the number of elements in the input repertoire. In conventional networks, the number of synapses needed is $M \times N$, which can climb very high when we have a large input repertoire, $M \gg N$.
Instead, manipulating ordinal patterns reduces the encoding space to $N$ synapses only. %The complexity of ordinal codes is therefore linear $O(N)$ and logarithmic for tree codes $O(log(N))$, but quadratic otherwise $O(N \times M)$. %We will develop this property in section~\ref{sec:roc_features} and in section~\ref{} when we compare this model to other neural architectures.
%
%This type of code is possible only if the size of chunks is known in advance. We can assume that ordinal codes of different length exist in a heterogeneous neural population to detect temporal location in sequences of different lengths.

Another way to code the ordinal structure of this time series is with a stack-ordered rule~\cite{Knutt1968} presented next in Fig.~\ref{fig:tree_code} c) and d), where the rank of each element is now computed with respect to the \emph{previously} stacked elements. That is, the first element becomes the central node from which the successive elements are organized around. New elements are compared iteratively with older ones and sorted one at a time relative to them. These stack-ordered sequences form binary search trees with the central node the first element of the sequence, with left and right branches respectively for lower and higher elements and deeper in the tree. The other elements in the sequence are stacked as sub-trees, preserving then the tree structure. We can retrieve the ordinal rank order of each element by reading from left to right each element organized in the tree and from top to bottom their depth in the sequence. 

%NEW
%For instance, element $18$ is the first element, and because the next element $13$ is lower, it will go to the successive left branch. Next element $8$ is compared with element $18$ and will go to the left branch as well. It is then compared with $13$ and will go as well to the left branch, as it is lower. Now, because the next element $14$ is lower than $18$ but higher than $13$, it will go to the left branch first then to the right between the two elements. Instead, because the next element $5$ is lower than $18$ and lower than $13$, it will go to the left branch first then again to the left. Finally, because the last element $19$ is bigger than $18$, it will go to the right branch.
%$seq:[18, 13, 8, 14, 5, 19]$
%NEW
%Despite the index is now lost, the resulting ordinal code gives some interval constraints about possible sequences and possible index within it; e.g., elements have a lower rank or a higher rank in comparison to others within a tree, see Fig.~\ref{fig:tree_code} b) and c).
%
%Thus, ordinal codes and tree-order codes can ease the task of retrieving elements by imposing interval constraints on the search space. 
%NEW

In computer science, binary search trees are well exploited to perform fast dichotomic search of elements in large ensembles by constraining progressively the search space of elements. Depending on the search method, the computational cost has been defined to be in linear polynomial time; the worst case corresponds to completely deployed trees in one branch, i.e. the rank-order codes.
Another property of stack-sortable permutation is to form a Dyck language, which is a string of balanced parentheses, capable to form any regular context-free grammar, arithmetic or algebraic expressions; see Fig.~\ref{fig:tree_code} e).
Thus, they can represent some regular expressions with particular hierarchical nested structure as found in language, logic, geometry, algebra or music~\cite{Dehaene2015}.

We propose to compare how the three different learning strategies encode the structural information in the synaptic weights of artificial neurons in order to be sensitive to the relative order, structure and hierarchies in sequences (Fig.~\ref{fig:tree_code} f).

\subsubsection{STDP learning mechanism} This first mechanism encodes the temporal order in the synaptic weights of one neuron~\cite{Izhikevich2004}; see the blue line in Fig.~\ref{fig:tree_code} f).
This Hebbian-based strategy encodes the temporal delays and order between elements and how close in time is one element to another.
One neuron can encode the particular sequence $seq:[18, 13, 8, 14, 5, 19]$ with respect to their temporal order in its weights $STDP\_weights:[1/6, 1/5, 1/4, 1/3, 1/2, 1]$, so that the later the element is in the sequence, the less sensitive it is to that element.
This strategy is not able to encode robustly one sequence as it looses easily accuracy over time due to the accumulated errors introduced by noise; i.e., the so-called vanishing gradient problem.
%
%LSTM networks are less prone to it due to attentional mechanism to keep one memory in time.
%Feed-forward (deep) networks, standard recurrent neural networks (with/out STDP) or hidden Markov models will easily loose accuracy after several iterations due to the accumulated errors because any errors, noise or, delays within a sequence and sensitivity to duration, will disrupt the sequence. LSTM networks are less prone to it due to attentional mechanism to keep one memory in time.

\subsubsection{Rank-order mechanism} This learning strategy instead can avoid this loss by encoding one sequence with respect to the relative order of its elements, in the synaptic weights of one neuron. The synaptic weights can follow the inverse ordinal ranking of the sequence $seq:[18, 13, 8, 14, 5, 19]$, which is $rank:[\#5, \#3, \#2, \#4, \#1, \#6]$, as follows $rank\_order\_weights:[1/2, 1/4, 1/5, 1/3, 1/6, 1/1]$; its weights are displayed with the red-dashed line in Fig.~\ref{fig:tree_code} f).

The last method corresponds to the tree-order strategy and implements a combination between a temporal-order code and an ordinal code to stack the elements iteratively in the binary tree, arranged with respect to their relative ranks. The synaptic weights follow the temporal order and the rank-order of each element in the sequence $seq:[18, 13, 8, 14, 5, 19]$, which is $tree\_order\_weights:[1/2, 1/4, 1/8, 1/16, 3/8, 3/4]$. The synaptic weights are represented by the green line in Fig.~\ref{fig:tree_code} f).

Ordinal trees might not be biologically plausible as they rely on the high precision encoding in the synaptic weights when the structure becomes more and more complex.
To overcome this problem, we present a fourth strategy in Fig.~\ref{fig:tree_code} f), a Huffman-like tree representation based on ordinal codes, which overcomes this limitation of tree codes. The ordinal codes replace the binary codes in the Huffman strategy, which is an optimal code in Information Theory for representing a sequence of discrete items. %The other advantages of this approach is that several ordinal codes of various lengths represent now the structural information of one tree, making this code more reliabe. 

\subsubsection{Ordinal STDP} 
\begin{figure*}[t!]
\begin{center}
\includegraphics[width=16cm]{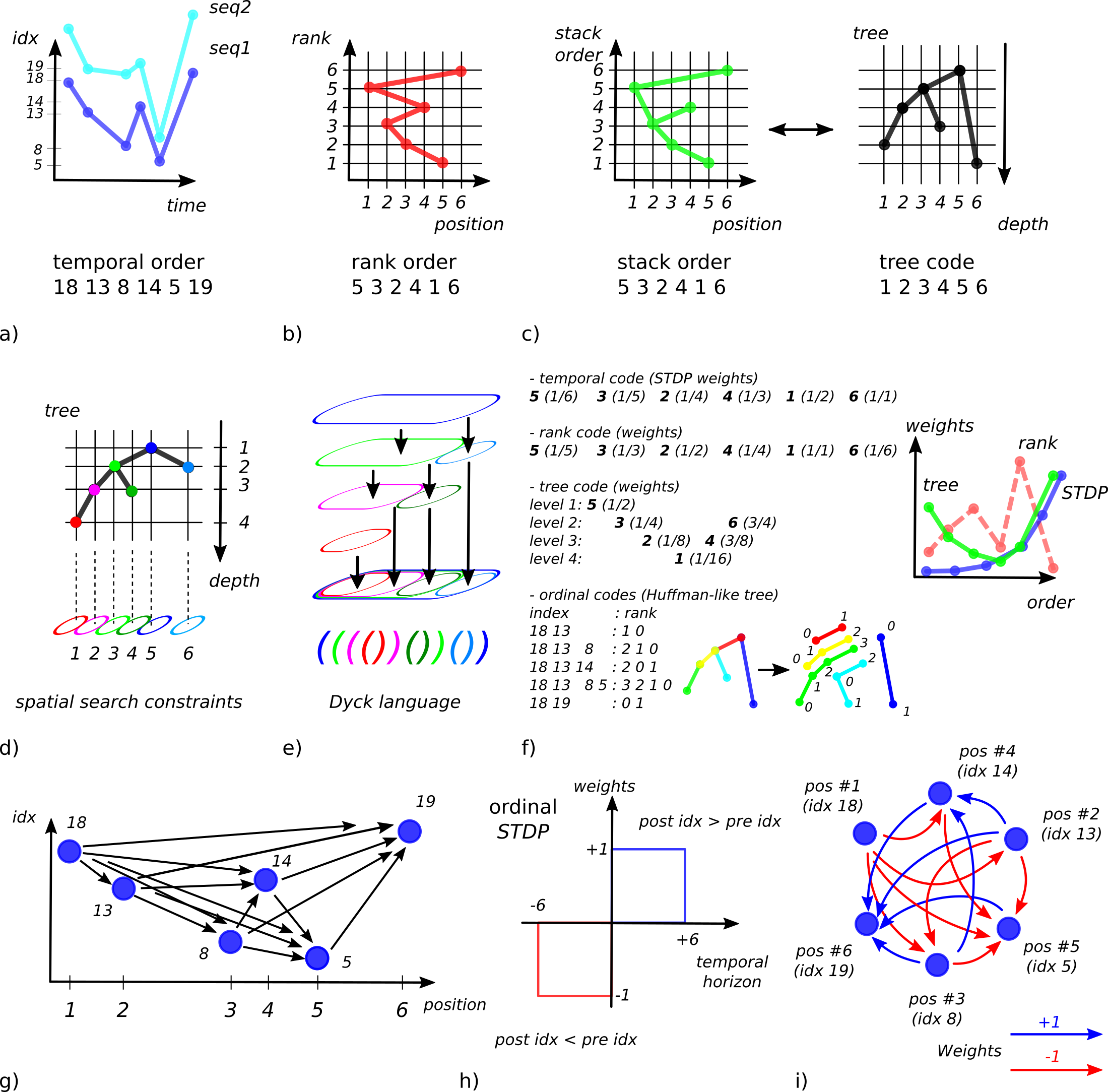}
\end{center}
\caption{Proposed neural models for representing the ordinal information in temporal sequences. We can describe the two-step process carried out with an ordinal  code and a tree-like ordinal code from the Spike Timing-dependent Plasticity rule. In a), a temporal sequence is displayed with different elements and index. In b), an ordinal code can serve to quantize a temporal sequence by suppressing the index of the neurons and by keeping only the relative order in it; i.e., their rank (\#) within the sequence. This process drastically reduces the amount of information to encode only its structure, irrespective of the neurons' index and their precise timing. In c), one different way to code the order in a sequence of items arriving iteratively is the stack-order code, which computes an ordinal code with respect to the previous elements already being stacked and ordered in it. Stack-ordered sequences form binary search trees with elements ordered both in time (depth) and in space (rank). In d), rank-order and tree-order codes can serve to represent hierarchical information and to retrieve or generate any new sequences following this latent structure as binary search trees in computer 's sorting and indexation tasks. In e), binary search trees form expressions in the Dyck language, which supports context-free grammars. The diagram in f) presents the synaptic weights for the different strategies to encode a sequence based on their temporal, rank or hierarchical (tree) order. A fourth way to encode ordinal information is with a Huffman-like tree representation with ordinal codes replacing binary digits for each branching. (g-h) present a fifth way to represent ordinal codes with an ordinal version of STDP. Pre- and post-synaptic connections between neurons encoding one specific ordinal location in a sequence can be wired with respect to the relative rank between two neurons: a negative (positive) weight for a negative (positive) index difference. In i), they can form an associative network with nodes sensitive only to ordinal information in sequences.}
\label{fig:tree_code}
\end{figure*}

it is an extension of the previous mechanism, presented in Figs.~\ref{fig:tree_code} (g-i).
This ordinal version of the STDP has a similar learning mechanism, expect that pre- and post-synaptic neurons reinforce their links with respect to their relative index, not their timing. Post-synaptic neurons with higher (resp. lower) index than pre-synaptic ones will strengthen (resp. diminish) their synaptic links. By doing so, spiking neurons sensitive to specific order in sequences can be constructed in recurrent networks, without the need to encode the indexes. In line with the principle of dynamic synchronization~\cite{Izhikevich2004}, this type of coding can be very robust to noise and the supression/insertion of nodes within the sequence.
It is also similar to a Hopfield network with neurons sensitive to ordinal information only, see Fig.~\ref{fig:tree_code} (i).

In the mathematical sense, ordinal codes are random permutations of integers. 
They have interesting computational and mathematical properties as they form a Dyck language and are used in combinatronix and lexicographic order for logic set theory, fast sorting, and expression-free grammar (such as LISP, PROLOG). As we attribute to the PFC some capabilities for symbol processing, which is a difficult task for neural networks, we think ordinal codes can help to achieve this process as it is found in classic AI.

Based on this argument, we propose that neurons implementing such kind of ordinal representation of information will possess per se all the necessary computational tools for encoding and decoding structure and hierarchy in sequences and for processing any type of language in all modalities (e.g., sound, action, vision).
This is to our knowledge novel and was never proposed before.

Since ordinal codes are sensitive to higher-order temporal patterns, they are affected nonlinearly by noise.
For instance, the encoding is very robust to local errors, but weak to structural error (noise on the rank order).
If the items 18 and 19 are inverted in the sequence [18, 13, 8, 14, 5, 19], the ordinal information will be swapped largely within it.
Therefore, the quantization makes the network to be attentive to the global structure in the signal and not to its details.
%
%Therefore ordinal codes are sensitive to this kind of swapping.

We think this feature is an important property for symbolic processing because the code [\#6, \#4, \#2, \#3, \#1, \#5] is strictly different to [\#5, \#3, \#2, \#4, \#1, \#6] in terms of structure. This can be advantageous for building a logical inference system as it is possible to validate or falsify incoming information from it.

%Furthermore, they may replace advantageously the convolution function in convolutional neural networks: because they are not sensitive to the particular index of the elements in a sequence but to the relative order only, ordinal codes are invariant to translation and to scale distortion as well, see Fig.~\ref{fig:tree_code} a).
%NEW
%Althought not necessarily named as fractal codes, tree-order and ordinal codes permit to represent information invariant to scales, because they are not sensitive to the particular index of the elements in a sequence but to their relative order, as seen in Fig.~\ref{fig:tree_code}.
%
%Self-similarity patterns can be advantageous for the organization of memory in neural networks, for compactness and for information retrieval.
%
%We will develop further these features for encoding and decoding tasks in section~\ref{sec:neurobio_encoding} and in section~\ref{sec:small_world}, in which we will discuss about the topology of brain networks in terms of information processing to organize memory.

% FIG 4 Tree Codes

\section{Discussions}
\label{Discussion}

We will present in section~\ref{Neuro} the neuro-cognitive foundations of our proposal and in section \ref{Cognition} the links to cognitive sciences and developmental studies.

\subsection{Neuro-cognitive foundations}
\label{Neuro}

We will detail in this section some phenomenons found in cognitive, developmental and brain sciences relying on the hierarchical representation of information for which we can give an interpretation within our framework. In section~\ref{sec:neurobio_encoding}, we suggest a functional role to the conjunctive cells and ordinal codes for the mechanism of nonlinear-mixed selectivity found in frontal cortex. In section~\ref{sec:broca_area}, we confront our ideas with brain observations on the neural mechanisms underlying language acquisition and the representation of syntactic structures. % in speech in the light of the ordinal and tree codes we propose.
In section~\ref{sec:mirror_neurons}, we explain how our framework can model higner-order representation of actions (the mirror neuron system). Finally in section~\ref{sec:small_world}, we develop our ideas on the functional organization of the brain and its development to construct hierarchical assemblies into one global memory workspace, the Connectome.

\subsubsection{Redundant nonlinear mixed selective codes}
\label{sec:neurobio_encoding}

The strongest argument in favor of our proposal comes from the discovery of very dynamic neurons found in the PFC~\cite{Romo1999, Machens2010, Tanji2001, Shima2007} functioning with conjunctive coding of multiple stimulus features, such as sensory stimuli, task rule, or motor response. 
One computational description of this phenomenon is done with the mechanism of nonlinear-mixed selectivity (NMS) proposed in~\cite{Rigotti2013, Fusi2016}. 
Within this framework, high-dimensional representations with mixed selectivity allow a simple linear readout to generate a huge number of different potential responses that depend on multiple task-relevant variables.

Recent studies have shown that the Lateral PFC hosts an abundance of these neurons with mixed selectivity~\cite{Parthasarathy2017, Sarma2016, Mansouri2006}. 
In particular, neurons with nonlinear mixed selectivity are thought to play a key role in the encoding of information~\cite{Fusi2016}. 
One recent paper explains further how it may be used by the brain to support reliable information transmission using unreliable neurons~\cite{Johnston577288}.

As a note, the parietal cortex possesses also conjunctive cells~\cite{Salinas2001, Andersen2002, Blohm2009, Genovesio2014}, that bind mutual information together for spatial transformation between different reference frames, multisensory alignment and decision making. 
Comparative neuroanatomical studies attribute similar functions to the parietal cortex and to the prefrontal cortex, representing relative metrics or conjunctive
representations~\cite{Genovesio2014} such as order with relative duration, and order with relative distance; but only the PFC is in a position to generate goal-based aims in context~\cite{Genovesio2009}. 
%
%Furthermore, neurons with mixed codes were also found in the Hippocampus and computational models for memory retention and regeneration were proposed to preserve robustly information using fractal codes~\cite{Tsuda2008, Yamaguti2011, Tsuda2015}, Haar/modulo codes [Gaussier unpublished], and generative models~\cite{Stoianov2018}.

%We would like to give some support on the NMS mechanism with additional hypothesis/features from our side. 
We would like to give new insights on the NMS mechanism based on our hypotheses. 
First, we see the role of the PFC neurons as much for encoding and categorization as for decoding and retrieving, which is not the case with the original formulation of the NMS mechanism.
Second, we suggest that what we might observe in their activity is their sensitivity to patterns, not to variables. Differently said, they might encode temporal or spatial structure without content (H1).
Third, the linear combination of mixed codes permits to represent efficiently nested representation of sequences, as showed in section~\ref{Model}. 
Thus, we suggest further that this linear combination of mixed codes can serve to retrieve efficiently sequences of neurons in very large assemblies based on their nonlinearity. In our framework, the nonlinearity effect is linked to the encoding of the ordinal rank of neural units from each other in one sequence (H2).

\subsubsection{Speech structure in the Broca area}
\label{sec:broca_area}

Since the seminal work of Broca, we know that the circuits in the left cortical hemisphere and in the prefrontal area implement language for perception and production of rule-based behaviors~\cite{Friederici2011} and can be regarded as the grammar center~\cite{Sakai2005, Ardila2011}.

Recently, it has been suggested that the Broca area plays a more general role as being a supra-modal ``Syntax Engine'' in the broaden sense, to abstract rules in other core domains and modalities such as music and action representation, as well as in visual scene understanding~\cite{Fogassi2007, Arbib2005, Arbib2008, Arbib2019, Fadiga2009, Gentilucci2006}.

%Although language has been long time thought separated from other types of cognition, syntactic rules and semantic exist in other domains as well (e.g., visual and motor) for which the Broca area plays also an integrative part of their underlying processing. 

The Broca area (Broadman area 44/45) interacts heavily with the Primary Auditory Cortex (PAC) and the Superior Temporal Gyrus (STG) located in the temporal area. The former is associated with the syntax in sentences and the later are associated with the sound representation.
Comparison of fMRI brain activation in sentence processing and nonlinguistic sequence mapping tasks~\cite{Hoen2006} found that 
BA44 was involved in both the processing of sentences and abstract structure in non-linguistic sequences whereas BA45 was exclusively activated in sentence processing~\cite{Arbib2014}. 

In speech processing, the Broca area is found sensitive to the order of events in sequences and its activity level is correlated with the syntactic complexity of the sentence~\cite{Dominey2003}.
For instance, the two slightly similar sentences \texttt{``Jean said \# Marie is great''} and \texttt{``Jean \# said Marie \# is great''} express different meaning and different relationships between the two persons.
They can be represented by different ordering trees for which the Broca area is sensitive with~\cite{Friederici2011}.
Experiments have shown that a higher complexity of the tree depth is correlated with a higher neural activity level in the Broca area region; see~\cite{Friederici2006a, Friederici2006b}. 

In line with~\cite{Dehaene2015, Udden2019, Fitch2012} who supports the view that the brain holds some exclusive mechanisms for manipulating symbolic nested trees, the Broca area appears clearly to hold one of those mechanisms for the detection of the complexity pattern in sequences~\cite{Koechlin2018}.
We might suspect that the Broca aera is functional very rapidly during infancy since babies and even neonates appear to be sensitive to syntax in proto-words~\cite{Saffran1996, Nazzi1998, Marcus1999, Gervain2008, Gervain2017}; see also the computational models of Dominey in~\cite{Dominey2000,Dominey2003, Dominey2006}.

In favor to this, recent experiments done in sentence processing by Meyer and colleagues present results showing the complementary work and the functional unity between the posterior cortical areas and the Broca area~\cite{Meyer2012}. Their experiment permitted to trace the linkage between the process of storage of verbs done in the posterior parts of the cortex and the processing of ordering elements done in the Broca area, the two working together.
According to our framework, we propose that the encoding done in the Broca area (H1) serves also for the decoding and the retrieving of the information stored in the posterior part of the cortex, and the type of code used represents ordinal tree codes (H2).

%In support of this separation between storage and ordering, experiments in text retrieval done in section~\ref{sec:experiments} showed how neurons sensitive to ordinal patterns (syntax) can be used for retrieving back elements in sequences (letters). We demonstrate in other computer simulations that a model of the Broca area could create ordinal representations of sequences in the sound domain as well~\cite{Pitti2020}; that is, sensitive to the temporal order of speech units composing them.
%
%Our neural simulation could extract easily hierarchical and syntax from sequences based on the rank-order coding and could generate (and retrieve) long sound sequences of fifty units length from the ordinal representations we selected.
%
%Because of the neural network encoded rank codes only, the computational cost was dramatically reduced and rapid in comparison with deep networks.\\

% Other neural implementations were proposed to resolve the manner how the frontal cortex extracts abstract structures in sequences. For instance, Dominey and colleagues proposed the Abstract Temporal Recurrent neural Network (ATRN) based on a different mechanism to model how the Broca area operates as a Short-Term Memory~\cite{Dominey2000, Dominey2003}.
% %
% Although ours might be simpler to implement and permits to reconstruct back the items' order in sequences, it provides a broader role played by structural information to organize, generate and retrieve back knowledge. \\

%FIG 5
\begin{figure*}[t!]
\begin{center}
\includegraphics[width=13cm]{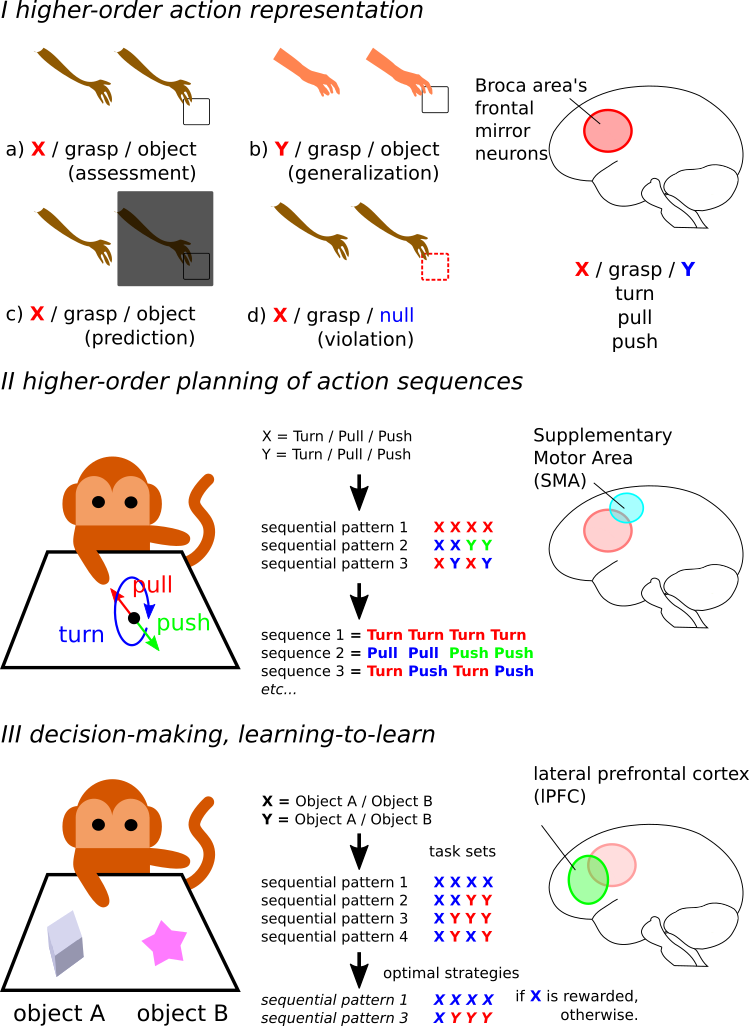}
\end{center}
\caption{Proposed neural mechanism for hierarchical planning and action representation in the Broca area and in the Mirror Neuron System (MNS). In $I$, our framework may provide some explanations on the results found for the functioning of the MNS. In $I-a$, mirror neurons are sensitive to higher-level action representation, the sequential ordering of action events, either self or other, if structural elements of the sequence are followed and respected; e.g., having one object to grasp. The violation of one of the structural elements in the sequence (e.g., not object to grasp) will not make the MN to fire. In $II$, neurons in the pre-supplementary motor area (pre-SMA) are sensitive to the higner-order of the action sequence (i.e., its 'grammar'). Tanji discovered in the pre-SMA of monkeys, a structure related to the Broca area in humans, the activation of neurons sensitive to specific sequential pattern but not to the lower-lever motor item constituting them~\cite{Tanji2007}. In $III$, Harlow's experiment on learning-to-learn, similar to Piaget's A-not-B problem~\cite{Smith1999}. Higher-order task sets can be learned to choose and infer the optimal strategy in a hierarchical reward-based decision making~\cite{Koechlin2018, Harlow1949} and for which reward conditioning on items does not work. An ordinal rule (a task set) can solve this problem.}
\label{fig:hierarchical_action}
\end{figure*}

\subsubsection{Representative motor schemata neurons, the mirror neurons}
\label{sec:mirror_neurons}

In support of a mechanism for detection and manipulation of highly structured information in the PFC other than speech, higher-level semantic action neurons called Mirror Neurons (MN) were found~\cite{Rizzolatti1998, Rizzolatti2004} in the pre-Supplementary Motor Area (SMA), see Fig.~\ref{fig:hierarchical_action}.
In these circuits, movement primitives coordinate to effect a wide variety of actions at a higher semantic-level to represent goal-directed motion such as grasping, holding, tiring, pulling etc... 
Mirror neurons have been discovered to fire for various action schemata depending on the type of grasp and goal~\cite{Rizzolatti1998, Arbib2005, Oztop2006, Oztop2013}. 
For instance, same mirror neurons fired whenever the monkey grasps an object from the left hand or even the right hand, and more surprisingly, when someone else also grasps an object, see Fig.~\ref{fig:hierarchical_action}-I.
%
%According to Arbib, 
This result supports the idea of representing neurons as a common substrate for motor preparation and imagery~\cite{Arbib2005, Arbib2008}.

%Arbib develops extensively a motor schema theory that we are in line with to explain Language by the assemblage of motor schemata or of representative neurons~\cite{Arbib1985, Arbib2005, Arbib2008}.
%
%Jeannerod describes particularly well his theory in~\cite{Jeannerod1994}: \emph{Arbib's view is that motor representations are composed of elementary schemas which are activated by object affordances and can adjust to visual input. During prehension, motor schemas for the subactions "reach," "preshape," "enclose," "rotate the forearm," or for selecting the number of fingers involved, would be available and would be selected automatically when required by object affordances}. 

In other experiments, Tanji and colleagues observed the sensitivity of SMA motor neurons to the structure of the motor sequence and not to the individual actions performed per se~\cite{Ninokura2004, Shima2007}. 
For instance, some of these higher-order neurons were sensitive to the motor pattern (AABB) or (ABAB) so that they could fire to any combinations of action primitives that follow this structure like Push-Push-Turn-Turn or Turn-Turn-Push-Push for the former and Push-Turn-Push-Turn or Turn-Tire-Turn-Tire for the later, as illustrated in Fig.~\ref{fig:hierarchical_action}-II.
Tanji proposed that these SMA neurons encode the structure in the motor sequence --, that is, the syntax of the task,-- but not its details (eg, the action unit) (H1).

%[reecrire] [sequence coding]
In similar experiments performed by Inoue and Mikami, some PFC neurons were found to modulate their amplitude level with respect to the position of items during the sequential presentation of two visual shape cues~\cite{Inoue2006}. The PFC neurons displayed graded activity with respect to their ordinal position within the sequence and to the visual shapes; e.g. first-ranked items, or second-ranked items (H1). In more complex tasks, PFC neurons were found to fire at particular moments within the sequence~\cite{Tanji2001}; e.g. the beginning, the middle, the end, or even throughout the evolution of the sequence. %Examples of visual tasks for which this ordinal information is important in cognitive development are presented in Fig.~\ref{fig:structural_learning} II and III and discussed in Appendix. % section~\ref{Cognitive}.

In line with this, Koechlin conducted experiments to isolate the functionalities of the Broca area and its implication to the hierarchical organization of human's behavior~\cite{Koechlin2006b}. He observed the formation of super-ordinate chunks based on the temporal structure of simpler actions. Accordingly, the Broca area processes hierarchical relations rather than cross-temporal contingencies between elements comprising action plans. Koechlin proposes that the Broca area is sensitive to the structural complexity of those action plans but insensitive to the variabilities of simple motor responses composing them (H1).
In a theoretical schematic model that he developed in~\cite{Koechlin2006b, Koechlin2007}, he describes how the Broca aera provides hierarchical control on lower regions in order to generate sequence of single acts (H2).

In so far, no valid theory can explain the development of the MNS, either genetic or epigenetics~\cite{Ferrari2013}. Associative models have been proposed~\cite{Heyes2010} althought they cannot account for goal encoding and action understanding. According to~\cite{Ferrari2013}, Hebbian processes --, in which repeated observations of self-produced actions are coupled with motor commands to create causal sensorimotor links,-- cannot explain as well the high variability and modulation found in the activity of mirror neurons. 

Our framework may provide a comprehensive understanding of the computational features underlying these \emph{representative} mirror neurons, in line with Simulation Theory~\cite{Jeannerod2001}, or of these ~\emph{super-ordinal} chunks, in line with Hierarchical Control based on Information Theory~\cite{Koechlin2007, Koechlin2016}. In our view, we may see MN as ``structure'' detector neurons insensitive to the raw action signals (H1), see Fig.~\ref{fig:hierarchical_action}. They may act then as ``template'' and ``variable fillers'' to generate (or simulate) any novel sequences with respect to the varying contexts: oneself-grasp or someone else, goal planning even in the dark, etc... (H2); see Fig.~\ref{fig:hierarchical_action} I. %Jeannerod1994
%
%Our theory resembles also to the dissociation processes for learning 'surface' structure and 'abstract' structure in sequences proposed by Dominey and colleagues in~\cite{Dominey1998}.
%
%Our proposal for ordinal coding and hierarchical neural trees can give some hints how neurons sensitive to higher-order semantics can develop for action understanding and language acquisition.\\

\subsubsection{Small-world network organization of cortical layers}
\label{sec:small_world}

% Using the metaphor of the mailman, the efficacy of the mailman to distribute letters depends not only on his own competence to deliver letters but also on how well the city is organized; e.g., well-ordered or randomly organized.
% %
% There are therefore some advantages to redesign cities at multiple hierarchical levels into districts, streets and building numbers for efficacy purpose; so might do the brain.

Many evidences suggest that the small-world organization of the different regions of the neocortex, the connectome, supports information processing within it~\cite{Sporns2004, Bassett2006, Friston2013}.
We suggest therefore the existence of hypotheses (H1) and (H2) contributing to (H3). 
%
%First, the well-ordering of the neocortex into hierarchies and small-world dynamics serve for fast and robust information retrieval of codes, which is not something clearly acknowledged in the computational neuroscience literature and no mechanism for cluster retrieval in complex systems is proposed in that sense. 
%
% For instance, in most researches, only the topological aspect of complex networks is emphasized such as the balanced level between segregation and integration~\cite{Varela2001, Sporns2004, Tognoli2014}, but the way how information is then retrieved in those networks is not examined.
% %
% Second, we propose that the mechanism for structure learning and decoding in the PFC may serve, as a second feature, to shape the well-ordering of the other parts of the brain for efficacy (H2). 
% %
% Detecting one structure in sequences gives the opportunity to generate sequences that possess also this particular structure, in an unsupervised manner.
% %
% Hence, we propose that the mechanisms for extracting structure in the world, performed by the PFC, serve for structuring memory networks in the brain itself, in line with the ideas of self-structuring information~\cite{Lungarella2005, Byrge2014}, neural scaffolding~\cite{Changeux1989a} and embrainment~\cite{McClelland2010}, see Fig.~\ref{fig:connectome} a-c). 

Although most neural models support the idea of bottom-up and local self-organization of random networks into complex small-world networks through synaptic connections and neuromodulation~\cite{Sporns2004, Heuvel2011}, we suggest that global synchrony via top-down and reentrant signals might play an important role as well for re-organizing information processing within networks for efficiency purpose in retrieval tasks, in line with top-down synchrony~\cite{Engel2001, Singer2003}, reentry~\cite{Tononi1992a, Tononi1992b} and the global working space~\cite{Dehaene1998, Dehaene2011}; this relates to our hypothesis (H3).
%as well  top-down self-organization as another important mechanism. , Engel2001b, 

In comparison to a random network, one organized network into small-world dynamics may favorish the faster retrieval of neurons and clusters.
Thus, we propose that the hierarchical rank-codes and tree-codes used as a first purpose for sequence retrieving in PFC (H2) may serve as a second purpose to control and to re-organize in a top-down manner the initial neural network for efficient decoding (H3).
Following  a developmental process, the hierarchical control in frontal area may shape the functional organization of the posterior parts of the neo-cortex, enhancing specialization and segregation (parcellation), see Fig.~\ref{fig:connectome} c).

\subsection{Links to Cognitive \& Developmental Sciences}
\label{Cognition}

% FIG 6 structural learning in development
\begin{figure*}[t!]
\begin{center}
\includegraphics[width=14cm]{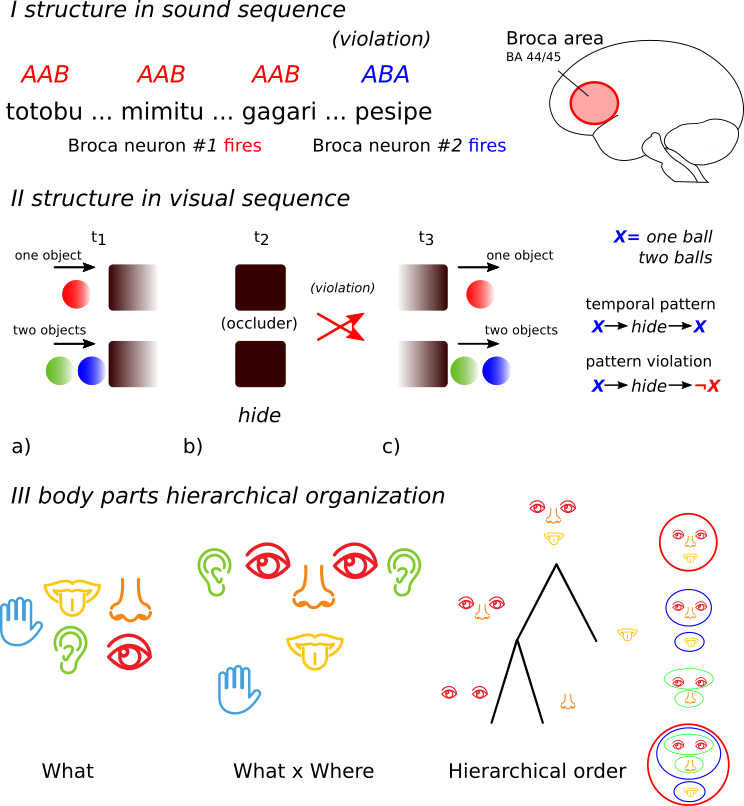}
\end{center}
\caption{Structure learning in core domains in infancy and linked to neurons in PFC sensitive to the rank order in sequence. Biological evidences and developmental data show that babies as early as 3 months-old of age are ready to grasp structure in data and to separate the hidden temporal pattern from the raw events. In sound processing in $I$, young infants were found to be sensitive to the structure of words like the examples above, independently to the syllables pronounced showing some activity in the left pre-frontal area4 (Broca area). In visual scene understanding in $II$, Baillargeon showed how infants were sensitive to the sequences coherency or to their violation. She argued that infants possess a Representational Engine to extract higher-level rules. Fadiga and colleagues in~\cite{Fadiga2009} showed similar results with the sensitivity of Broca area neurons sensitive to the order in visual sequences. In $III$, hypothesis of the body parts hierarchical organization and parcellation. Our framework can propose a neural implementation how hierarchical information can be constructed in the Broca area and in the human Mirror Neuron System. The tree codes model we proposed can serve to represent the What (items) and Where (their location) information in images, and their hierarchies order in the tree (e.g. a face). Higher order representation can help to recognize rapidly one shape; e.g. for facial imation. It can serve as well for the top-down parcellation of brain networks to represent the body; i.e. the body image.}
\label{fig:structural_learning}
\end{figure*}

% \subsection{Links to Cognitive Sciences \& Developmental studies}
% \label{Cognitive}

Several developmental observations suggest that babies at birth are capable to detect structured information and to manipulate symbolic representations as proposed by~\cite{Gopnik2004}.
%
%cannot perform causal inference, and manipulate abstract knwoledge .
%
For instance, early in development, infants are keen on grasping structure in core domains~\cite{Spelke2003, Spelke2007}, inferring causal models and making hypotheses on problems about space, time, numerosity, language and psychology~\cite{Gopnik2000, Tenenbaum2011, Baillargeon2012}. 
%
%They rapidly develop knowledge about the structure of the world about space, time and psychology but it is only at around 8 months that they gain the aptitudes to make complex sequences and to retain structural information in their environment.

%
%Current AI systems are still poorly endowed with such capabilities because they rely heavily on statistical learning and cannot perform causal inference, and manipulate abstract knwoledge as proposed by~\cite{Gopnik2004, Meltzoff2007a}.
%
We propose that if infants are particularly good at extracting structural information from raw data, it is due to the mechanisms found in the PFC responsible for gathering structural information in signals and for manipulating it at an abstract-level (H1).

We will develop this idea how infants detect higher-level information in visual scenes in section~\ref{sec:visual_scenes}, on the emergence of language and syntactic capabilities in babies in section~\ref{sec:protowords} and to the resolving of higher-cognitive tasks in section~\ref{sec:learning-to-learn}. Finally, we will extend our reflexions how our framework contributes to a novel view of embodied cognition in section~\ref{sec:agency}.

\subsubsection{Structure understanding in visual scenes}
\label{sec:visual_scenes}

In scene understanding tasks, Baillargeon discovered that young infants reason at an abstract level about objects' categories (inanimated, animated), physics (occlusion, shadows, rigid 3D objects do not distort or disappear) and properties (soft, round) to simulate what will happen next.
She proposes that a link exists between language and event representations~\cite{Baillargeon1994, Baillargeon2012} and that infants possess a 'Physical Reasoning System' to endow them with a grasp of intuitive physics about objects.
%
%According to Baillargeon, infants may possess a Physical Reasoning System to endow them with a grasp of intuitive physics about objects.

Similar to grammar in language, this Physical Reasoning System may allow infants to reason about causal and physical events and to detect several physical violations occuring in the visual scene. 

For instance, infants were surprised when an object placed behind a screen disappeared and that a different object reappeared after. Young infants possess therefore some expectation about physical knowledge, such as object permanence and object occlusion.

As illustrated in Figure~\ref{fig:structural_learning}-II for object permanence, we present a coherent scenario with either one or two objects occluded by a screen and reappearing after form a coherent sequence with the same structure ABA and possible endings (object 1$\rightarrow$hide$\rightarrow$object 1 and object 2$\rightarrow$hide$\rightarrow$object 2). Another scenario with two impossible endings (object 1$\rightarrow$hide$\rightarrow$object 2 and object 2$\rightarrow$hide$\rightarrow$object 1) will conduct to a violation of the learned rule ABA.

In order to solve this task, some developmental scientists have suggested that infants have to reason at a super-ordinal level of representation (e.g., the structural or symbolic level) and not at the raw pixel level as classical Machine Learning algorithms would usually do~\cite{Gopnik2017}.
The Bayesian theory and probabilistic inference are often taken as good candidates to explain infant behaviors for these tasks but we think they are not enough to explain how fast learning is done and how generalization from few examples is carried on~\cite{Tenenbaum2006, Tenenbaum2011, Gopnik2017}.
For instance, Bayesian networks are directed graphes difficult to use when it comes to manipulate the density probabilities and the causal relationships of a large amount of data and high numer of classes. Furthermore, the framework of Bayesian theory cannot explain how the compositionality of a rule can be done so readily and how the acquisition of novel symbols can be managed fom raw data.
Bayesian nets avoid as well how symbol grounding is done. %, how the probability density terms are estimated and how the correct graph is selected for belief propagation.

In comparison, our framework may explain how the frontal neurons can extract the ordinal structure very rapidly in sequences (H1). Thus, we suggest that extracting the super-ordinal representations in sequences is instead a very rapid mechanism, which can be used readily to infer the structure of an unseen sequence or to generate new samples based on it. %this structure as well as doing it very rapidly.
This differs from the classic Bayesian approach in the sense that one system does not need many trials to learn probability distributions and conditional probabilities to generate new sequences.

In computational models, extracting relational information with rank-order or tree-order codes in raw sequences permits to represent temporal and hierarchical structures and to have expectations on future events in visual scenes.
Furthermore, since rank-order codes can produce context-free grammars as proposed in section~II-A, we can potentially define rules and construct logical systems with them. %~\ref{sec:roc_features}
% Furthermore, since rank-order codes can produce context-free grammars as proposed in section~\ref{sec:roc_features}, we can potentially define rules and construct logical systems with them.

Therefore, in line with the Physical Reasoning System proposed by Baillargeon, we propose that infants have access to a ``visual language'' thanks to the processing performed in the PFC for extracting structures and rules in visual sequences.
Such system is in anycase the implementation of a physical simulator at the pixel level as proposed in~\cite{lake2017}, but more the collection of structural patterns that permit to falsify logically relational information across incoming data (rule-based reasoning and fact checking), to anticipate future events (causal inference) or to generate one visual planning (compositionality and imagination).

\subsubsection{Extracting structure in proto-words}
\label{sec:protowords}

In language processing, seminal works by Saffran~\cite{Saffran1996,Saffran2003}, Marcus~\cite{Marcus1999, Marcus2007}, Nazzi~\cite{Nazzi1998, Nazzi2019} and Gervain~\cite{Gervain2008,Gervain2017} have showed that babies after 8 months and even neonates are capable to learn artificial grammars and to extract structure in proto-words, like the AAB pattern in the words 'totobu', 'gagari', 'mimitu', although they were not familiar with the specific temporal order of the sound sequence. They showed also that they were sensitive to structure violation if other patterns were presented such as the word 'pesipe' with the ABA pattern~\cite{Nazzi2019, DehaeneLambertz2014, Dehaene2015}; see Figure~\ref{fig:structural_learning}-I. 

% Despite the immaturity of the baby's brain, its performances in this task indicate that a neural mechanism is at work in order to grasp statistical regularities and structure within speech sequences to conform on a set of grammatical rules that are learned.
%
% Interestingly, the PFC has been found active during these tasks as well as the Broca area. 

The mechanism we propose to extract the ordinal information in sequences irrespective to the inputs identity may explain the processes behind these results found in babies and neonates and how the PFC and Broca area is potentially performing it. %~\cite{Pitti2020}.
This is in line with models of the frontal areas that extract abstract structural information in temporal sequences in~\cite{Dominey2000, Dominey2003} as well as our computational experiments in speech structure in~\cite{Pitti2020}. % or the experiment done previously on text structure.

% Similarly, our computational experiments done with the neural architecture Inferno Gate permitted to extract temporal structure in sequences of 50 items over a large sound repertoire of 14.000 items, to detect temporal structure violation and to generate easily novel sequences following one temporal order~\cite{Pitti2020}. Such capabilities were not possible to perform by the current state-of-art recurrent network LSTM.
%
%Our proposal is not the first attempt to explain these results and Dominey proposed one neural network that could extract temporal patterns in sequences also.
%
%However, the extraction mechanism based on bio-inspired model of the rank-order coding, the computational cost and the organisation of information that we propose is more computationally efficient and in similar to a gating mechanism

Recent developmental and ethological comparisons defend the idea that the emergence of language comes from the functional maturation of the human infant brain and does not have a physiological cause~\cite{Boe2019} as supported by the current dominant theory~\cite{Lieberman1968}. %; e.g. the different configuration of the larynx vocal system in monkeys and babies, which is still .
It has been suggested that conjunctive cells in frontal areas play an important role for goal-based behaviors~\cite{Genovesio2009, Genovesio2014}.
We suggest further that hierarchical tree codes and rank-order codes may allow the structural learning of tree representations in temporal sequences and that they are necessary for grammar and language~\cite{Fitch2012, Udden2019}.

\subsubsection{Hierarchical structure in higher-level cognitive tasks and planning}
\label{sec:learning-to-learn}

%%%%
We can note two experiments employed to emphasize the importance of the frontal areas for problem-solving in visual scenes using task sets~\cite{Koechlin2016, Koechlin2018}, one from Harlow experiment on task sets~\cite{Harlow1942, Harlow1949} and the other from Piaget experiment on the A-not-B error test~\cite{Diamond1985, Diamond1989a}.
The two experiments play on repetition and novelty, on masking and on the use of temporal delayed information to predict either the spatial location in the case of the A-not-B error experiment or of the temporal strategy to employ in the case of Harlow's Task Sets.
Importantly, none of the two tests can be explained by simple conditioning and associative reinforcement learning. And both experiments are considered as steps-tones of infant cognitive development. % to access  in infant cognition to their role in higher cognition.
%Adolph2005

Recent simulations of the frontal areas could successfully model these experiments using neural fields~\cite{McClelland2010}, LSTM~\cite{Wang2018} or spiking neurons~\cite{Pitti2013b}.
%Smith1999
These models were based on the active maintenance of information or the inhibition of spurious one.
However, since the frontal areas extract hierarchical structure in temporal sequences, we propose different explanations and models to explain higher-order planning in cognitive tasks, see Fig.~\ref{fig:hierarchical_action} III.

First, one observation that we can make is that the Piagetian experiment of the A-not-B test is relatively similar to the experiments performed by Baillargeon discussed in section~\ref{sec:visual_scenes} and in Fig.~\ref{fig:structural_learning}-II except that one object remains hidden in one location A or B and is uncovered after a delay.
The ordinal structure XYX is still respected, except that here the child has to predict the spatial location of the hidden object, A or B, which corresponds to the last term in the sequence XY[X], X replaced by the A or B location.

Second, in Harlow experiment, one successful strategy for the monkey is to open door A or door B and to learn that the same reward will be given if he chose for the next five trials the same location irrespective to the input stimulus provided.
One potential explanation to understand this experiment based on our framework is that the monkey has learned two strategies XXXXXX or XYYYYY, and depending if he receives one reward or not on the first trial irrespective to the door location, X=A or X=B, the first or second strategy will be selected on the second trial, see Fig.~\ref{fig:hierarchical_action} III.

%current theories behind emphasized the role played by the prefrontal cortex to form a working memory to delays to select and inhib
% 
% the working memory and learning  model-based done in to related to 
% [Harlow, task set et A non B]
% Marcus express that the connexi
% In so far, there is some parallel with the test proposed by 

\subsubsection{multimodal binding and supra-modal correspondences}
\label{sec:agency}

We can reinterpret certain experiments and theories in cognitive sciences on enaction and embodiment, in the light of our framework and of the functioning of the PFC presented in section~IV-A1. %\ref{Neuro}. % and hypotheisis may permit to interpret differently certain experiments and data found in the cognitive literature. 
For instance, since the Broca area encodes super-ordinal information independent to raw input as described in section~IV-A2 and in section~IV-A3, and starts functioning early at birth, it might make sense that the detection of higher-order structure in sensorimotor raw signals is possible even for babies. 
The mechanisms and the hypotheses we proposed on the detection of higher-order events and on memory organization can support the view that babies may possess the tools to manipulate higher-order information and knowledge at birth.

For instance, we propose that the theories based on supra-modal integration by Meltzoff and colleagues~\cite{Meltzoff1997, Marshall2014, Meltzoff2018} to explain facial imitation and body representation or the amodal 'contingency' detection proposed by O'Regan and colleagues~\cite{ORegan2001, Terekhov2016} to explain perception as an amodal process, can be supported by the information processing we described in the PFC.
We will describe below how our model can explain the different theories we described above, for extracting hierarchical knowledge from raw sensorimotor signals. % based on the rank-order and tree-order representations of information.
%
%We make the note that the word 'contingency' is ambiguous as it can be misinterpreted as a low-level synchronization process between sensing and acting or between neurons.

These higher-order representations may help to create this \emph{lingua franca} in the brain that makes to interpret and connect the different modalities from each others~\cite{Guellai2019}. % or to interpret self and others.
For instance, O'Regan and Noe~\cite{ORegan2001} explain their theory of sensorimotor coordination to associate the \emph{structure of the changes} between raw actions and sensory modalities, if they follow similar temporal patterns.
This theory is different with Hebbian learning, which associates one element with another but not their relative changes, which is henceforth amodal.
In that sense, the relative changes between multiple modalities correspond to an amodal temporal structure extracted from these signals. %, generating a similar sequence in the missing modality.
%
%Accordingly, an amodal information on observational learning may serve the learning of physical causal events and also the observational learning in the social domain.
%
%In our view, the correspondence between modalities can be possible through the detection of similar spatio-temporal structures: 
%As an example, the ascending pattern $ABCD$ can be equivalent either to a sound pitch augmenting in crescendo, or to the light luminosity augmenting in intensity or to the hands moving to one direction in accordance.
%
%Thus, the property of the PFC to extract hierarchical structure from raw information may \emph{calibrate} the body signals from each others at an abstract-level, in the sense given by O'Regan and Noe~\cite{ORegan2001}, to associate the structure of the changes between raw actions and sensory modalities, if they follow similar temporal patterns.

Terekhov and O'Regan explain how the amodal notions of space and geometry can be abstractly extracted from relative information about sensorimotor signals and by constructing an amodal function of it~\cite{Terekhov2016}.
Accordingly, other notions taken in core domains and difficult to define can be expressed in the same way; e.g., numerosity, acceleration, mass, gravity or softness. %Those qualitative properties of perceptual experiences correspond to the theory of the Qualia, which is also linked to the theory of integrated information~\cite{Balduzzi2009}.
%
%The mechanism we propose to represent amodal information in time series may provide a neural explanation to this theory of the sensorimotor approach to perception~\cite{ORegan2001}, which emphasizes the central role of information transformations in perception~\cite{ORegan2011}, instead of information mapping and representation. 

% For instance, most models of sensorimotor integration and agency are based on contingency detection and are seen as low-level processes, via a comparator model~\cite{Watson1966, Watson1994, Hiraki2006}. 
% %
% However, sensorimotor integration and agency can be seen also as a top-down and amodal process via the detection of the same superordinal structure between two or more modalities and the generation of the corresponding sequence in another modality with same pattern. Hence the structure detection in frontal areas may govern the agentive hierarchical control of posterior areas and  the supervision also of imitative behaviors.
% %
% We propose that sensitivity to the relation between events in the visual domain may permit to generate a similar spatio-temporal sequence in the motor domain in a plastic way based on the mechanism found in the Broca area.
% 
% Sensorimotor contingency detection appears very early in infancy~\cite{Nadel2005, Hiraki2006}. 

A theory in line also with the supra-modal representation system we proposed is the ones of Meltzoff in~\cite{Meltzoff1997, Meltzoff2018} with the ``Like-Me'' theory and the 'active intermodal mapping' (AIM) account of early imitation.
AIM proposes that an intermodal comparison is possible because the perception and production of human acts are represented within a common (supramodal) framework. %This common framework is proposed to be based on organ identification~\cite{} : ``the newborns' first response to seeing a particular facial gesture is activation of the corresponding body part. (...) It is as if young infants isolate what part of their body to move before how to move it. We call this 'organ identification'.''
%
%Furthermore, this theory is not directly based on associative Hebbian learning as ``early imitation is not entirely stimulus bound, directly triggered, or reflexive''. Accordingly, imitation is goal-directed and generative whose aim is 'matching the target'.
%
The recent experiments done by Marshall and Meltzoff associated to the frontal mirror neuron system~\cite{Marshall2014} could confort their theory as the Mu rhythm originating from this area to the somatosensory area has been found active during motor babbling and during body parts parcellation and identification. % appear induced from it.

In figure~\ref{fig:structural_learning}-III, we present a supra-modal representation of facial elements, hierarchically ordered as our framework proposed it. Each element (eye, nose, mouth, etc... or the WHAT) is represented relative to others in space and in a structured fashion (the WHERE). This higher-level tree representation that we presume constructed in the PFC can serve then for detecting the global structure of our own face (WHAT $\times$ WHERE), then for mapping the body parts into the somatosensory area. % as proposed by Meltzoff and then for identifying the organs of other faces by matching the higher order structure, elements free.
%
%An outstanding question would be if such a supramodal processing is at work at birth for all modalities, giving rise to this notion of agency and of a bodily Self.
%
% Experiments on the Mu rhythm activation and cancellation associated to the body babbling and originating in the somatosensory area and in the mirror neurons system~\cite{Marshall2014}, may give some arguments in favor of the idea that an agentivity process is at work very early~\cite{Rochat2001, Rochat2003, Nadel2005, Rochat2019}.
%
%Results for which our framework is in line with.
%Meltzoff Learning to make things happen: Infants’ observational learning of social and physical causal events

\section{Conclusion}

In order to keep trace of information, the brain has to deal with the problems of canceling intrinsic noise and of harnessing its own complexity.
It has to resolve the problem to locate where information is and how to index new ones.
The neural mechanism in charge of this task has to be capable to manipulate neural addresses and to map the brain's own neural circuitry; its connectome.
Such tool is important for information processing and preservation, but also for memory formation and retrieval.

As a shift in evolution, we propose that the neural mechanism used by the PFC to detect structure in temporal sequences, based on the temporal order of incoming information, serves as second purpose to the spatial ordering and indexing of brain networks.
This top-down self-organization is done for efficacy.

The sensitivity of PFC units to serial order can serve for incoming events as for patterning the brain dynamics. % can permit to produce a compact code representing the spatial location of neurons distributed over different places, but part of the same temporal cluster.
At the unit level, ordinal codes lose information about the exact location of neurons (their address).
At the population level, however, ordinal codes may retrieve it using redundancy.
Since they manipulate relative addresses, ordinal codes may act as neuronal pointers and the type of information processing they are doing may be seen as \emph{digital}.%, similar to Fourier transform.
%
%
%In this line of thought, there should be a sufficient number of ordinal codes to reconstruct perfectly the original memory sequence,%, which would satisfy the Nyquist-Shannon sampling theorem.
%The Nyquist–Shannon sampling theorem is a theorem in the field of digital signal processing which serves as a fundamental bridge between continuous-time signals and discrete-time signals.It establishes a sufficient condition for a sample rate that permits a discrete sequence of samples to capture all the information from a continuous-time signal of finite bandwidth
%
These ideas are in line with recent proposals that the PFC is the brain router that may manipulate neuronal variables and pointers to construct a neuronal global workspace for conscious access~\cite{Zylberberg2010, Dehaene2011} or others that the brain manipulates integrated and differentiated information codes~\cite{Balduzzi2009}.

\section*{Acknowledgements}

AP would like to dedicate this manuscript in memory of Chalom Pitti (1943-2019).

\bibliographystyle{IEEEtran}
\bibliography{references}

\end{document}